\documentclass[lettersize,journal]{IEEEtran}
\usepackage{amsmath,amsfonts}
\usepackage{algorithmic}
\usepackage{algorithm}
\usepackage{array}
\usepackage[caption=false,font=normalsize,labelfont=sf,textfont=sf]{subfig}
\usepackage{textcomp}
\usepackage{stfloats}
\usepackage{url}
\usepackage{verbatim}
\usepackage{graphicx}
\usepackage{cite}
\usepackage{balance}
\usepackage{xcolor}
\definecolor{green}{HTML}{228B22}
\newcommand\red[1]{\textcolor{red}{#1}}
\newcommand\green[1]{\textcolor{green}{#1}}

\usepackage{etoolbox}
\makeatletter
\patchcmd{\@makecaption}
  {\scshape}
  {}
  {}
  {}
\makeatother

\begin{document}

\title{Are Current Task-oriented Dialogue Systems Able to Satisfy Impolite Users?}

\author{
\IEEEauthorblockN{Zhiqiang Hu\IEEEauthorrefmark{1}, Roy Ka-Wei Lee\IEEEauthorrefmark{1}, Nancy F. Chen\IEEEauthorrefmark{2}}\\
% \vspace{0.05in}
\IEEEauthorblockA{\IEEEauthorrefmark{1}Singapore University of Technology and Design, Singapore \\ \emph{zhiqiang\_hu@mymail.sutd.edu.sg, roy\_lee@sutd.edu.sg}} \\
\IEEEauthorblockA{\IEEEauthorrefmark{2}Institute of Infocomm Research (I2R), A*STAR, Singapore \\ \emph{nfychen@i2r.a-star.edu.sg}}
%\vspace{0.05in}
        % <-this % stops a space
% \thanks{This paper was produced by the IEEE Publication Technology Group. They are in Piscataway, NJ.}% <-this % stops a space
% \thanks{Manuscript received April 19, 2021; revised August 16, 2021.}
}

% The paper headers
% \markboth{Journal of \LaTeX\ Class Files,~Vol.~14, No.~8, August~2021}%
% {Zhiqiang \MakeLowercase{\textit{et al.}}: Are Current Task-oriented Dialogue Systems Able to Satisfy Impolite Users?}

%\IEEEpubid{0000--0000/00\$00.00~\copyright~2021 IEEE}
% Remember, if you use this you must call \IEEEpubidadjcol in the second
% column for its text to clear the IEEEpubid mark.

\maketitle

\begin{abstract}
Task-oriented dialogue (TOD) systems have assisted users on many tasks, including ticket booking and service inquiries. While existing TOD systems have shown promising performance in serving customer needs, these systems mostly assume that users would interact with the dialogue agent politely. This assumption is unrealistic as impatient or frustrated customers may also interact with TOD systems impolitely. This paper aims to address this research gap by investigating impolite users' effects on TOD systems. Specifically, we constructed an impolite dialogue corpus and conducted extensive experiments to evaluate the state-of-the-art TOD systems on our impolite dialogue corpus. Our experimental results show that existing TOD systems are unable to handle impolite user utterances. We also present a data augmentation method to improve TOD performance in impolite dialogues. Nevertheless, handling impolite dialogues remains a very challenging research task. We hope by releasing the impolite dialogue corpus and establishing the benchmark evaluations, more researchers are encouraged to investigate this new challenging research task. 
\end{abstract}


\section{Introduction}
\label{sec:intro}
\textbf{Motivation.} Task-oriented dialogue (TOD) systems play a vital role in many businesses and service operations. Specifically, these systems are deployed to assist users with specific tasks such as ticket booking and restaurant reservations through natural language conversations. TOD systems are usually built through a pipeline architecture that consists of four sequential modules, including natural language understanding (NLU), dialogue state tracking (DST), policy learning (POL), and natural language generation (NLG) \cite{elder2020make,balakrishnan2019constrained,li2020slot,golovanov2019large}. More recently, researchers have also explored leveraging large pre-trained language models to improve the performance of TOD systems \cite{Yang2021UBARTF,hosseini2020simple,lin-etal-2020-mintl}. These TOD systems have demonstrated their effectiveness in understanding and responding to the users' needs through conversations. 

As most TOD systems are developed to serve and assist humans in performing specific tasks, the politeness of the TD systems remains a key design consideration. For instance, Gupta et al. \cite{gupta2007rude} presented POLLy (Politeness for Language Learning), a system that combines a spoken language generator with an AI Planner to model Brown and Levinson's theory of politeness in TOD. Bothe et al. \cite{bothe2018towards}  developed a dialogue-based navigation approach incorporating politeness and sociolinguistic features for robotic behavioral modeling. More recently, Mishra et al. \cite{MISHRA2022242} proposed a politeness adaptive dialogue system (PADS) that can interact with users politely and showcases empathy. 

\begin{figure}[t] 
	\centering
	\includegraphics[scale = 0.65]{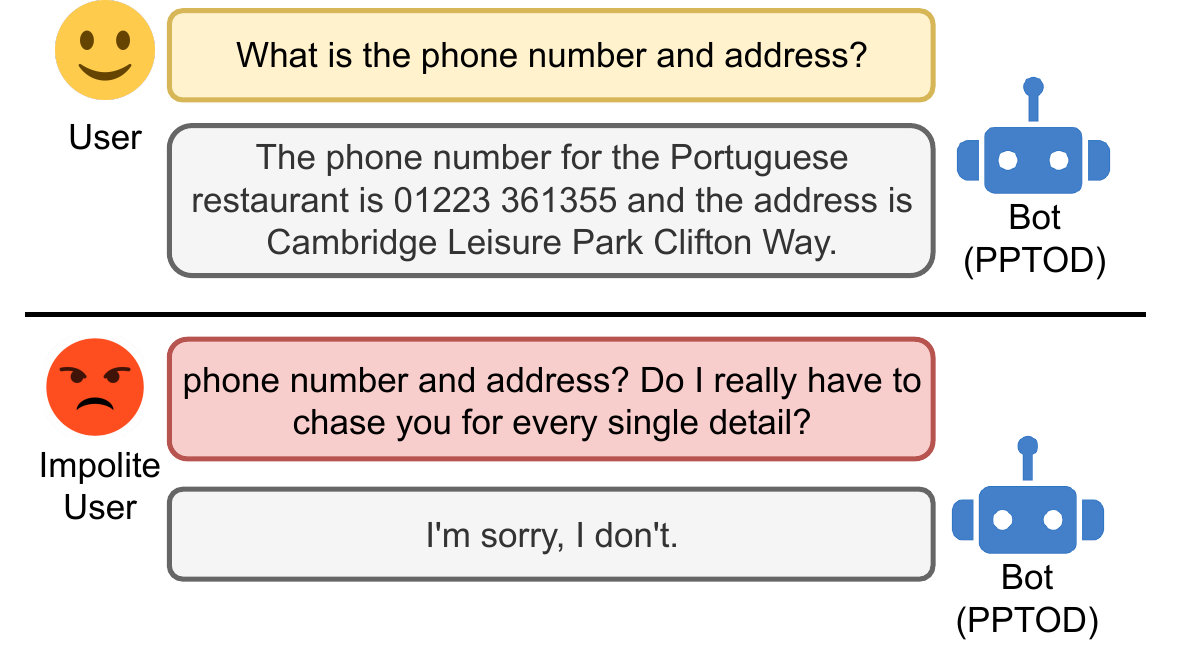} 
	\caption{Examples of two dialogue interactions between PPTOD and two types of users: normal user (top) and impolite user (bottom).}
	\label{fig:example}
\end{figure}

Nevertheless, the above studies have focused on generating polite dialogues and ignored the users' politeness (or impoliteness) in the conversation. Therefore it is unclear how the TOD systems would respond when the users interact with the TOD systems in an impolite manner, especially when the users are in a rush to get information or frustrated when TOD systems provide irrelevant responses. Consider the example in Figure~\ref{fig:example}, we notice that the TOD system, PPOTD \cite{su2021multi}, is able to provide a proper response to a user who presents the question in a normal or polite manner. However, when encountering an impolite user, PPOTD is not able to provide a proper and relevant response. Ideally, the TOD systems should be robust in handling user requests regardless of their politeness.

A straightforward approach to improving TOD systems' ability to handle impolite users is to train the dialogue systems with impolite user utterances. Unfortunately, most of the existing TOD datasets \cite{budzianowski-etal-2018-multiwoz,zang-etal-2020-multiwoz,rastogi2020towards,zhu2020crosswoz} only capture user utterances that are neural or polite. The lack of an impolite dialogue dataset also limits the evaluation of TOD systems; to the best of our knowledge, there are no existing studies on the robustness of TOD systems in handling problematic users.

%This is problematic as TOD systems deployed in business settings should be able to handle users' requests regardless of their politeness or tone.

\textbf{Research Objectives.} To address the research gaps, we aim to investigate the effects of impolite user utterances on TOD systems. Working towards this goal, we collect and annotate an impolite dialogue corpus by manually rewriting the user utterances of the MultiWOZ 2.2 dataset \cite{zang-etal-2020-multiwoz}. Specifically, human annotators are recruited to rewrite the user utterances with role-playing scenarios that could encourage impolite user utterances. For example, ``\textit{imagine you are in a rush and frustrated that the systems have given the wrong response for the second time.}''. In total, the human annotators have rewritten over 10K impolite user utterances. Statistical and linguistic analyses of the impolite user utterances are also performed to understand the constructed dataset better. 

The impolite dialogue corpus is subsequently used to evaluate the performance and limitations of state-of-the-art TOD systems. Specifically, we have designed experiments to evaluate the robustness of TOD systems in handling impolite users and understand the effects of impolite user utterances on these systems. We have also explored solutions to improve TOD systems' robustness in handling impolite users. A possible solution is to train the TOD systems with more data. However, the construction of a large-scale impolite dialogue corpus is a laborious and expensive process. Therefore, we propose a data augmentation method that utilizes text style transfer techniques to improve TOD systems' performance in impolite dialogues. 

\textbf{Contributions.} We summarize our contributions as follows: 

\begin{itemize}
    \item We collect and annotate an impolite dialogue corpus to support the evaluation of TOD systems performance when handling impolite users. We hope that the impolite dialogue dataset will encourage researchers to propose TOD systems that are robust in handling users' requests.
    \item We evaluate the performance of six state-of-the-art TOD systems using our impolite dialogue corpus. The evaluation results showed that existing TOD systems have difficulty handling impolite users' requests.
    \item We propose a simple data augmentation method that utilizes text style transfer techniques to improve TOD systems' performance in impolite dialogues. 
\end{itemize}

\section{Related Work}
\label{sec:related}

\subsubsection{Task Oriented Dialogue Systems}
With the rapid advancement of deep learning techniques, TOD systems have shown promising performance in handling user requests and interactions. Recent studies have focused on end-to-end TOD systems to train a general mapping from user utterance to the system's natural language response~\cite{lee-2021-improving-end,Zhang2020TaskOrientedDS, zhang-etal-2020-probabilistic,bordes2016learning,lei-etal-2018-sequicity}. Yang et al. \cite{Yang2021UBARTF} proposed UBAR by fine-tuning the large pre-trained unidirectional language model GPT-2 \cite{radford2019language} on the entire dialog session sequence. The dialogue session consists of user utterances, belief states, database results, system actions, and system responses of every dialog turn. Su et al. \cite{su2021multi} proposed PPTOD to effectively leverage pre-trained language models with a multi-task pre-training strategy that increases the model's ability with heterogeneous dialogue corpora. Lin et al. \cite{lin-etal-2020-mintl} proposed Minimalist Transfer Learning (MinTL) to plug-and-play large-scale pre-trained models for domain transfer in dialogue task completion. Zang et al. \cite{zang-etal-2020-multiwoz} proposed the LABES model, which treated the dialogue states as discrete latent variables to reduce the reliance on turn-level DST labels. Kulhánek et al. \cite{kulhanek-etal-2021-augpt} proposed AuGPT with modified training objectives for language model fine-tuning and data augmentation via back-translation \cite{edunov-etal-2018-understanding} to increase the diversity of the training data. Existing studies have also leveraged knowledge bases to track pivotal and critical information required in generating TOD system agent's responses \cite{zhu2017flexible,eric2017key,ghazvininejad2018knowledge,hua2020learning}. For instance, Madotto et al. \cite{madotto2020learning} dynamically updated a knowledge base via fine-tuning by directly embedding it into the model parameters. Other studies have also explored reinforcement learning to build TOD systems \cite{zhao2016towards,li2016user,liu2017iterative}. For instance, Zhao et al. \cite{zhao2016towards} utilized a Deep Recurrent Q-Networks (DRQN) for building TOD systems.

\subsubsection{Modeling Politeness in Dialogue Systems}
Recent studies have also attempted to improve dialogue systems to generate responses in a more empathetic manner \cite{ma2020survey,shi-yu-2018-sentiment,feng2021emowoz,li2020interactive}. Yu et al. \cite{shi-yu-2018-sentiment} proposed to include user sentiment obtained through multimodal information (acoustic, dialogic, and textual) in the end-to-end learning framework to make TOD systems more user-adaptive and effective. Feng et al. \cite{feng2021emowoz} constructed a corpus containing task-oriented dialogues with emotion labels for emotion recognition in TOD systems. However, the impoliteness of users is not modeled as too few instances exist in the MultiWOZ dataset. The lack of impolite dialogue data motivates us to construct an impolite dialogue corpus to facilitate downstream analysis.

Politeness is a human virtue and a crucial aspect of communication~\cite{brown1978universals,danescu-niculescu-mizil-etal-2013-computational}. Danescu-Niculescu-Mizil et al. \cite{danescu-niculescu-mizil-etal-2013-computational} proposed a computational framework to identify the linguistic aspects of politeness with application to social factors. Researchers have also attempted to model and include politeness in TOD systems~\cite{golchha2019courteously,niu2018polite,madaan2020politeness}. For instance, Golchha et al. \cite{golchha2019courteously} utilized a reinforced pointer generator network to transform a generic response into a polite response. More recently, Madaan et al. \cite{madaan2020politeness} adopted a text style transfer approach to generate polite sentences while preserving the intended content. Nevertheless, most of these studies have focused on generating polite responses, neglecting the handling of impolite inputs, i.e., impolite user utterances. This study aims to fill this research gap by extensively evaluating state-of-the-art TOD systems' ability to handle impolite dialogues.

\subsubsection{Data Augmentation in Dialogue Systems}

Data augmentation, which aims to enlarge training data size in machine learning systems, is a common solution to the data scarcity problem. Data augmentation methods has also been widely used in dialogue systems \cite{kulhanek-etal-2021-augpt,hou-etal-2018-sequence,gritta-etal-2021-conversation}. For instance, Kurata et al. \cite{kurata2016labeled} trained an encoder-decoder to reconstruct the utterances in training data. To augment training data, the encoder's output hidden states are perturbed randomly to yield different utterances. Hou et al. \cite{hou-etal-2018-sequence} proposed a sequence-to-sequence generation-based data augmentation framework that models relations between utterances of the same semantic frame in the training data. Gritta et al. \cite{gritta-etal-2021-conversation} proposed the Conversation Graph (ConvGraph), which is a graph-based representation of dialogues, to augment data volume and diversity by generating dialogue paths. In this paper, we propose a simple data augmentation method that utilizes text style transfer techniques to generate impolite user utterances for training data to improve TOD systems' performance in dealing with impolite users.

\section{Impolite Dialogue Corpus}
\label{sec:data}
We construct an impolite dialogue dataset to support our evaluation of TOD systems' ability to interpret and respond to impolite users. Specifically, we recruited eight native English speakers to rewrite the user utterances in MultiWOZ 2.2 dataset~\cite{zang-etal-2020-multiwoz} in an impolite manner. To the best of our knowledge, this is the first impolite task-oriented dialogue corpus. In the subsequent sections, we will discuss the corpus construction process and provide a preliminary analysis of the constructed impolite dialogue corpus.

\subsection{Corpus Construction}
\label{data_construction}
MultiWOZ 2.2 \cite{zang-etal-2020-multiwoz} is a large-scale multi-domain task-oriented dialogue benchmark that contains dialogues in seven domains, including attraction, hotel, hospital, bus, restaurant, train, and taxi. This dataset is also popular and commonly used to evaluate existing TOD systems \cite{Zhang2020TaskOrientedDS,zhang-etal-2020-probabilistic,lin-etal-2020-mintl,Yang2021UBARTF,su2021multi,kulhanek-etal-2021-augpt}. We performed a preliminary analysis using the Stanford Politeness classifier trained on Wikipedia requests data \cite{danescu-niculescu-mizil-etal-2013-computational} to assign a politeness score to the user utterances in MultiWOZ 2.2. We found that 99\% of the user utterances are classified as \textit{polite}. Hence, we aim to rewrite the user utterance in MultiWOZ 2.2 to create our impolite dialogue corpus. 

\textbf{Impolite Rewriting.} The goal is to rewrite the user utterance in the MultiWOZ 2.2 dataset and present the user utterance in a rude and impolite manner. We randomly sampled a subset of dialogues from MultiWOZ 2.2 for rewriting. Next, we recruited eight native English speakers to rewrite the user utterances. For each user utterance, the annotators are presented with the entire dialogue history to have the conversation's overall context. The annotators are tasked to rewrite the user utterances with three objectives: (i) the rewritten sentences should be impolite, (ii) the content of the rewritten sentence should be semantically close to the original sentence, and (iii) the rewritten sentences should be fluent. To further encourage the diversity of the impolite user utterance, we also prescribed six role-playing scenarios to aid the annotators in the rewriting tasks. For instance, we asked the annotators to imagine they were customers in a bad mood or impatient customers who wanted to get the information quickly. The details of the role-playing scenarios are shown in Table~\ref{tbl:scenario}, and the annotation system interface is included in the Appendix~\ref{apx:tool_interface}. 

% \subsection{Scenarios for Impolite User Annotation}
% \label{apx:scenarios}
%In consideration of real-life situations, we predefined six scenarios shown in Table \ref{tbl:scenario}. Annotators are asked to transfer user utterances to impolite utterances based on one of the scenarios to improve the diversity of impolite user data.

\begin{table}[htb]
\small
\centering
\caption{Role-playing scenarios for impolite user annotation.}
\label{tbl:scenario}
\begin{tabular}{cp{6.5cm}}
\hline
\textbf{No.} & \textbf{Scenario} \\
\hline\hline
1 & Imagine that the customer is a sarcastic person in a bad mood. \\
2 & Imagine that the customer is an impatient customer that wants to get the information fast. \\
3 & Imagine that the customer is in a bad mood as something bad has just happened (e.g., just had an argument with friends or spouse). \\
4 & Imagine that the customer is tired and hungry after a long-haul flight and need to get this information fast. \\
5 & Imagine that the customer is a spoilt-brat with a lot of money. \\
6 & Imagine that the customer is getting help from CSA for the third time and they didn't get the previous information right. \\\hline

\end{tabular}
\end{table}

\textbf{Annotation Quality Control.} Impoliteness is subjective, and the annotators may have different interpretations of impoliteness. Therefore, we implement iterative checkpoints to evaluate the quality of the rewritten user utterance. Specifically, we conducted peer evaluation at various checkpoints to allow annotators to rate the quality of each other's rewritten sentences. The annotators are tasked to rate the rewritten user utterance based on the following three criteria:

\begin{itemize}
    \item \textit{Politeness.} Rate the sentence politeness using a 5-point Likert scale. 1: strongly opined that the sentence is impolite, and 5: strongly opined that the sentence is polite.
    \item \textit{Content Preservation.} Compare the original user utterance and rewrite sentence and rate the amount of content preserved in the rewrite sentence using a 5-point Likert scale. 1: the original and rewritten sentences have very different content, and 5: the original and rewritten sentences have the same content.
    \item \textit{Fluency.} Rate the fluency of rewritten sentences using a 5-point Likert scale. 1: unreadable with too many grammatical errors, 5: perfectly fluent sentence.
\end{itemize}

Each user utterance is evaluated by two annotators. Nevertheless, we recognize that it is unnatural for all utterances in a dialogue to be impolite. Thus, we would consider a dialogue impolite if 50\% of the user utterances in a conversation are rated as impolite (i.e., with \textit{Politeness} score 2 or less). This exercise allows the annotators to align their understanding of the rewriting task. The annotators will revise the unqualified dialogues until they are rated impolite in the peer evaluation. While the annotators are tasked to assess each other’s work, they are unaware of their evaluation scores to mitigate any biases. Therefore, the annotators might learn new ways to write impolite dialogues from each other, but they are not writing to “\textit{optimize}” any assessment scores in the human evaluation.

\begin{table*}[t]
\small
\centering
\caption{Domain distribution of MultiWOZ 2.2 and our annotated impolite dialogue corpus. Numbers in () represent the percentage of dialogues in each domain.}
\label{tbl:domian_distribution}
\begin{tabular}{c|ccccccc}
\hline
\textbf{Data} & \textbf{Restaurant} & \textbf{Attraction} & \textbf{Hotel} & \textbf{Taxi} & \textbf{Train\&Bus} & \textbf{Hospital} \\
\hline\hline
MultiWOZ 2.2 & 3836 (45.5\%) & 2681 (31.8\%) & 3369 (39.9\%) & 1463 (17.3\%) & 2969 (35.2\%) &  107 (1.3\%)\\
Impolite Dialogue Corpus & 694 (44.0\%) & 545 (28.8\%) & 630 (40.0\%) & 295 (18.7\%) & 528 (33.5\%) & 107 (6.8\%) \\\hline

\end{tabular}
\end{table*}

\subsection{Corpus Analysis}
\label{data_analysis}
In total, the annotators rewrote 1,573 dialogues, comprising 10,667 user utterances. Table~\ref{tbl:domian_distribution} shows the distributions of the MultiWOZ 2.2 dataset and our impolite dialogue corpus. As we have sampled a substantial number of dialogues from MultiWOZ 2.20, we notice that the rewritten impolite dialogues follow similar domain distributions as the original dataset.

\begin{table}[t]
\small
\centering
\caption{Peer evaluation results.}
\label{tbl:peer_evaluation}
\begin{tabular}{c|c}
\hline
\textbf{Metric} & \textbf{Avg. Score} \\
\hline\hline
Politeness & 1.96 \\
Content Preservation & 4.68 \\
Fluency & 4.67 \\\hline
\end{tabular}
\end{table}

\begin{figure}[t] 
	\centering
	\includegraphics[scale = 0.465]{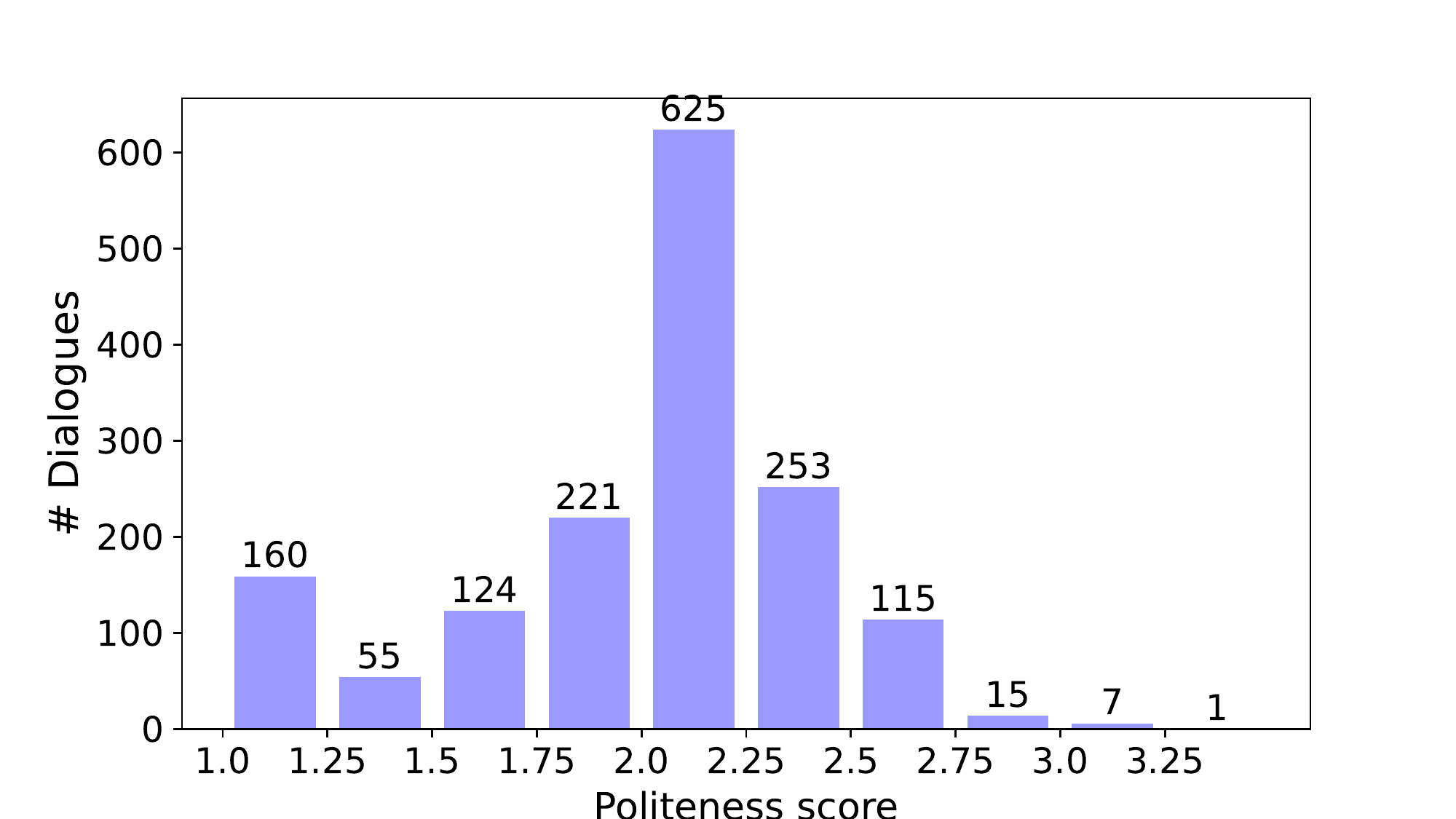} 
	\caption{Distribution of dialogues binned according to the politeness scores.}
	\label{fig:politeness_score}
\end{figure}

Table \ref{tbl:peer_evaluation} shows the results of the final peer evaluation of all rewritten impolite user utterances. Specifically, the average \textit{politeness}, \textit{content preservation}, and \textit{fluency} scores of the rewritten impolite user utterances are reported. The high average \textit{content preservation} and \textit{fluency} scores suggest that the high-quality rewritten utterances retain the original user's intention in the conversations. More importantly, the average \textit{politeness} score is 1.96, indicating that most of the rewrites are impolite but not too ``\textit{offensive}''. We further examine and show the \textit{politeness} score distribution of the rewritten impolite dialogue in Figure~\ref{fig:politeness_score}. Note that the \textit{politeness} score of dialogue is obtained by averaging the \textit{politeness} scores of the rewritten user utterances in the given dialogue. We observe that most rewritten dialogues are impolite, having \textit{politeness} scores of less than 2.5. 

We also empirically examine the top 20 keywords in the original and rewritten dialogues (shown in Table~\ref{tbl:keyword}. We notice that in the original dialogues, users tend to express gratitude by using terms such as ``\textit{thank}'' and adopt courtesy terms such ``\textit{please}'', ``\textit{help}''. Users in original dialogues also posted their requests as questions using terms such as ``\textit{would you}''. Conversely, in the rewritten dialogues, gratitude terms are absent. Users are less courteous and issued direct commands to the agent using terms such as ``\textit{find}'', ``\textit{get}'', ``\textit{give}''. Frequent terms ``\textit{hurry}'' and ``\textit{fast}'' also suggested the users' impatience in the dialogues.  

As we aim to keep the rewritten utterances natural and realistic, we did not limit the users to using offensive language (neither did we encourage them). Interestingly, we have also checked the impolite user utterances and found offensive languages (e.g., ``\textit{idiot}'', ``\textit{stupid}'', ``\textit{f*ck}'', etc.) being used in some of the rewritten utterances.

\begin{table}[t]
\small
\centering
\caption{Top 10 keywords in original and rewritten dialogues.}
\label{tbl:keyword}
\begin{tabular}{p{3.5cm}|p{3.5cm}}
\hline
\textbf{Top 20 keywords in Original Dialogues} & \textbf{Top 20 keywords in Rewritten Dialogues} \\
\hline\hline
need, \green{please}, \green{thank}, yes, like, looking, \green{would}, number, book, also, restaurant, \green{thanks}, \green{help}, train, hotel, Cambridge, people, \red{find}, free, \red{get} & \red{find}, \red{get}, \red{give}, time, need, book, want, number, go, \red{hurry}, ok, \red{fast}, one, restaurant,\red{make}, people, job, better, train, Cambridge \\\hline
\end{tabular}
\end{table}

\section{Models}
\label{sec:model}

We experiment with six state-of-the-art TOD systems and evaluate their performance in handling user utterances in our constructed impolite dialogue dataset. These TOD systems are not designed to respond to impolite users or trained with impolite dialogues. Thus, they may not perform well on our corpus. A simple approach to improve the TOD systems' performance is to train them with impolite user utterances. However, there is inadequate impolite dialogue data to train these systems. Therefore, we propose a data augmentation strategy to enhance the TOD systems' ability to handle impolite users. 

%~\cite{Yang2021UBARTF,Zhang2020TaskOrientedDS,zhang-etal-2020-probabilistic,lin-etal-2020-mintl,kulhanek-etal-2021-augpt,su2021multi}

%To evaluate the ability of current TOD systems to handle impolite users, we examined six state-of-the-art E2E TOD models \cite{Yang2021UBARTF,Zhang2020TaskOrientedDS,zhang-etal-2020-probabilistic,lin-etal-2020-mintl,kulhanek-etal-2021-augpt,su2021multi} on the impolite user dialogues. Due to the limited ability of current TOD models, we propose a direct and simple data augmentation method which is based on text style transfer techniques to improve the performance of these models on impolite users. We summarize the models and our data augmentation method below and direct the reader to supplementary material for further detail.

\subsection{TOD Systems}
Recent studies have proposed TOD systems with promising performance on the MultiWOZ 2.2 dataset. For our study, we select six TOD systems that have achieved state-of-the-art performance on the task-oriented dialogue generation task. Domain Aware Multi-Decoder (\textbf{DAMD})~\cite{Zhang2020TaskOrientedDS} utilizes the \textit{one-to-many} property that assumes that multiple responses may be appropriate for the same dialog context to generate diverse dialogue responses. \textbf{LABES}~ \cite{zhang-etal-2020-probabilistic} is a probabilistic dialogue model with belief states represented as discrete latent variables and jointly modeled with system responses. \textbf{MinTL}~\cite{lin-etal-2020-mintl} is a transfer learning framework that allows plug-and-play pre-trained seq2seq models to jointly learn dialogue state tracking and dialogue response generation. \textbf{UBAR}~\cite{Yang2021UBARTF} fine-tunes GPT-2~\cite{radford2019language} on the entire dialogue session sequence for dialogue response generation. Similarly, \textbf{AuGPT}~\cite{kulhanek-etal-2021-augpt} fine-tunes GPT-2 with modified training objectives and enhanced data through back-translation. \textbf{PPTOD}~\cite{su2021multi} unifies the TOD task as multiple generation tasks, including intent detection, dialogue state tracking, and response generation.

%More recent work proposed E2E TOD systems which achieved good performance in their respective tasks, with the help of recent relevant advances. We choose six E2E TOD models that achieve state-of-the-art performance on the MultiWOZ 2.2 dataset. Domain Aware Multi-Decoder (\textbf{DAMD}) \cite{Zhang2020TaskOrientedDS} utilizes the one-to-many property that multiple responses can be appropriate for the same dialog context, to generate diverse appropriate dialog responses. \textbf{LABES} \cite{zhang-etal-2020-probabilistic} is a probabilistic dialog model with belief states represented as discrete latent variables and jointly modeled with system responses. \textbf{MinTL} \cite{lin-etal-2020-mintl} is a transfer learning framework that allows plug-and-play pre-trained seq2seq models and jointly learn dialogue state tracking and dialogue response generation. \textbf{UBAR} \cite{Yang2021UBARTF} fine-tunes GPT-2 \cite{radford2019language} on the entire dialogue session sequence. Similarly, \textbf{AuGPT} \cite{kulhanek-etal-2021-augpt} fine-tunes GPT-2 with modified training objectives and enhanced data through back-translation. \textbf{PPTOD} \cite{su2021multi} unifies the TOD task as multiple generation tasks, including intent detection, DST, and response generation.

\subsection{Data Augmentation with Text Style Transfer}
Training TOD systems with impolite dialogues can improve their capabilities in handling impolite users. However, collecting impolite dialogues is a challenging task as human annotation is a laborious and expensive process. Therefore, we explore augmenting TOD systems by generating impolite user utterances automatically. We formulate the impolite utterance generation as a supervised text style transfer task \cite{hu2010text} (i.e., politeness transfer). Specifically, the text style transfer algorithms will be trained using a parallel dataset; we have pairs of aligned polite (i.e., original) and the corresponding impolite (i.e., rewritten) user utterances in our impolite dialogue corpus. Subsequently, the trained text style transfer algorithms will be able to take in unseen user utterances from MultiWOZ 2.2 dataset as input and generate impolite user utterances as output. Finally, the generated impolite user utterances will be augmented to train TOD systems.   

In our implementation, we first fine-tune the pre-trained language model \textit{T5-large} \cite{JMLR:v21:20-074} and BART-large \cite{lewis-etal-2020-bart} with polite user utterances as inputs and impolite rewrites as targets. Then, the fine-tuned T5 and BART model is used to transfer the politeness for user utterances in the rest of the dialogues in MultiWOZ 2.2 dataset. We have also included a state-of-the-art text style transfer model, DAST \cite{li-etal-2019-domain}, to perform the politeness transfer task. 

We evaluate the text style transferred user utterance on three automatic metrics, including \textit{politeness}, \textit{content preservation}, and \textit{fluency}. For \textit{politeness}, we train a politeness classifier based on the BERT base model \cite{devlin2018bert} with the original polite user utterances and corresponding impolite rewrites. The trained classifier predicts if a given user utterance is correctly transferred to \textit{impolite} style, and the accuracy of the predictions is reported. For\textit{ content preservation}, we compute the BLEU score \cite{papineni-etal-2002-bleu} between the transferred sentences and original user utterances. For \textit{fluency}, we use GPT-2 \cite{radford2019language} to calculate perplexity score (PPL) on the transferred sentences. Finally, We compute the geometric mean (G-Mean score) of ACC, BLEU, and 1/PPL to give an overall score of the models' performance. We take the inverse
of the calculated perplexity score because a lower PPL score corresponds to better fluency. These evaluation metrics are commonly used in existing text style transfer studies \cite{hu2022text}.

Table \ref{tbl:tst_results} shows the automatic evaluation results of the style transferred user utterances. We observe that both pre-trained language models have achieved reasonably good performance on the politeness style transfer task. Specifically, the fine-tuned T5-large model achieves the best performance on the G-Mean score and 85.3\%. The transferred user utterances are mostly impolite (i.e., high accuracy score) and fluent (i.e., low PPL). The content is also well-presented. Interestingly, we noted that DAST did not perform well for the politeness transfer task. A possible reason could be that the DAST is designed to perform non-parallel text style transfer, and the model did not exploit the parallel information in the training data~\cite{li-etal-2019-domain}. Another reason could be the small training dataset; DAST is trained on our impolite dialogue corpus, while the T5 and BART are pre-trained with a larger corpus and fine-tuned on our impolite dialogue dataset.

The automatically generated impolite user utterances from the fine-tuned T5-Large are augmented to the original MultiWOZ 2.2 dataset to train the TOD systems. We will discuss the effects of data augmentation in our experiment section.

%A straightforward idea to improve the performance of the TOD models mentioned above on rude users is to add impolite user data to the training set. However, human annotation is a laborious and expensive process. We formulate the generation of impolite user utterances as a supervised text style transfer task (politeness transfer), because the alignment information of polite and impolite user utterances can be obtained from our collected data. We fine-tune the pre-trained language model T5 large \cite{JMLR:v21:20-074} with polite user utterances as inputs and impolite rewrites as targets. Then, the fine-tuned T5 model is used to transfer the politeness for user utterances of rest dialogues in MultiWOZ 2.2 dataset. We evaluate the generated sentences on three automatic metrics, including politeness, content preservation, and fluency. Specifically, for politeness, we train a politeness classifier based on the BERT base model \cite{devlin2018bert} with the original polite user utterances and respective impolite rewrites. For content preservation, we employ the BLEU score \cite{papineni-etal-2002-bleu} between generated sentences and original user utterances. For fluency, we use GPT-2 \cite{radford2019language} to calculate perplexity score (PPL) on the generated sentences. Table \ref{tbl:tst_results} shows the automatic evaluation results of augmented data. We observe that 85.3\% of the original user utterances have been transferred to impolite. The content has been well preserved, and the generated sentences are fluent, according to BLEU score 43.4 and PPL score 6.3.

\begin{table}[t]
\small
\centering
\caption{Automatic evaluation results of text style transfer task.}
\label{tbl:tst_results}
\begin{tabular}{ccccc}
\hline
\textbf{Model} & \textbf{ACC}(\%) & \textbf{BLEU} & \textbf{PPL} & \textbf{G-Mean}\\
\hline\hline
DAST & 73.6 & 25.4 & 28.5 & 4.03 \\
BART-large Fine-tune & \textbf{96.4} & 37.6 & \textbf{6.2} & 8.36 \\
T5-large Fine-tune & 85.3 & \textbf{43.4} & 6.3 & \textbf{8.38}\\
\hline
\end{tabular}
\end{table}

\section{Evaluation}
\label{sec:evaluation}

One of the primary goals of this study is to evaluate the TOD systems' ability to handle impolite dialogues. Therefore, we design automatic and human evaluation experiments to benchmark the performance of TOD systems on our impolite dialogue corpus.

%We conduct a thorough and extensive evaluation utilizing both humans and automatic metrics to quantify the performance of current TOD models on impolite users. 

\subsection{Automatic Evaluation}
Over the past decades, many different automatic evaluation methods \cite{sun2021simulating,peng2020raddle,deriu2021survey} have been proposed to evaluate TOD systems. The evaluation methodologies are inextricably linked to the properties of the evaluated dialogue system. For instance, Nekvinda et al. \cite{nekvinda2021shades} proposed their standalone standardized evaluation scripts for the MultiWOZ dataset to eliminate inconsistencies in data preprocessing and reporting of evaluation metrics, i.e., BLEU score and Inform \& Success rates. For our study, we employ four automatic evaluation metrics that are commonly used in existing studies~\cite{budzianowski-etal-2018-multiwoz,nekvinda2021shades} to benchmark the performance of TOD systems on impolite dialogues:

\begin{itemize}
    \item \textit{Inform}: The inform rate is the proportion of dialogues in which the system mentions a name of an entity that does not conflict with the current dialogue state or the user's goal.
    \item \textit{Success}: The percentage of dialogues in which the system provides the correct entity and answers all the requested information.
    \item \textit{BLEU}: The BLEU score between the generated utterances and the ground truth is used to approximate the output fluency.
    \item \textit{Combined}. The combined score is computed through $(Inform + Success) \times 0.5 + BLEU$ as an overall quality measure suggested in \cite{mehri2019structured}.
\end{itemize}

\subsection{Human Evaluation}
There are relatively lesser studies that performed a human evaluation on TOD systems \cite{Zhang2020TaskOrientedDS,kulhanek-etal-2021-augpt,su2021multi} as such evaluations are often expensive and laborious. In our study, we perform a human evaluation to access the TOD systems on three criteria: The human evaluation is conducted on a random subset of 100 dialogues with 50 of the original dialogue in the MultiWOZ 2.2 dataset and 50 corresponding dialogues from our impolite dialogue corpus. We recruit four human evaluators, and at least two human evaluators evaluate each dialogue. The human evaluators are tasked to evaluate the generated dialogues on the following criteria:  

\begin{itemize}
    \item \textit{Success}: A binary indicator (yes or no) on whether the dialogue system fulfills the information requirements dictated by the user's goals. For instance, this includes whether the dialogue system has found the correct type of venue and whether the dialogue system returned all the requested information. 
    \item \textit{Comprehension}: A 5-point Likert scale that indicates the TOD system's level of comprehension of the user input. 1: the TOD system did not understand the user intention; 5: TOD system understands well the user input. 
    \item \textit{Appropriateness}: A 5-point Likert scale indicates if the TOD system's response is appropriate and human-like. 1: the TOD system's response is inappropriate and does not make sense; 5: TOD system's response is appropriate and human-like. 
\end{itemize}

\section{Experiments}
\label{sec:experiment}

In this section, we design experiments to evaluate the state-of-the-art TOD systems' ability to handle impolite users using our impolite dialogue corpus. The rest of this section is organized as follows: We first present the details of the experimental settings. Next, we perform automatic and human evaluations of the TOD systems and analyze their performance on our impolite dialogue corpus. Finally, we conduct further empirical analyses to understand the TOD systems' challenges in handling impolite users.

%In this section, we describe the experiments conducted to evaluate the state-of-the-art TOD systems on our impolite dialogue corpus. We first present the details of reproducing the TOD systems. Next, we discuss the results from automatic and human evaluations. Finally, we showcase some case studies of the generated responses for impolite dialogue and analyze the difficulty of handling impolite user. 

%To evaluate the ability of state-of-the-art TOD system to handle impolite users, we conduct extensive experiments with six TOD systems on the collected impolite user dialogues and corresponding original dialogues. In this section, we first introduce the details of reproducing and retraining with augmented data for every TOD system. Then, we discuss the performance and limitations of TOD systems based on the results of automatic and human evaluation. Finally, we discuss the correlation between the difficulty of handling impolite users and different factors, such as politeness, sentence length, and semantic similarity.

\subsection{Experimental Settings}
\textbf{Reproduce Models.} We reproduce six state-of-the-art TOD systems by training them on the MultiWOZ 2.2 dataset. For TOD systems that have publicly released checkpoints, such as AuGPT~\cite{kulhanek-etal-2021-augpt} and PPTOD~\cite{su2021multi}, we directly use the released checkpoints trained on  MultiWOZ 2.2. For DAMD~\cite{Zhang2020TaskOrientedDS}, UBAR~\cite{Yang2021UBARTF}, MinTL~\cite{lin-etal-2020-mintl}, and LABES~\cite{zhang-etal-2020-probabilistic}, we use their published code implementations and optimize their hyperparameters to reproduce their reported results for MultiWOZ 2.2 dataset\footnote{https://github.com/budzianowski/MultiWOZ}. Table \ref{tbl:reproduce_results} shows the reported and reproduced performance of the TOD systems tested on the MultiWOZ 2.2 dataset. We observe that the reproduced performance is equivalent to or slightly better than the reported results. Our subsequent experiment will evaluate the reproduced TOD systems on the impolite dialogue corpus.

\textbf{Data Augmentation.} To evaluate the effectiveness of our proposed data augmentation solution, we train the six TOD systems with MultiWOZ 2.2 dataset augmented with the automatically generated impolite dialogues. The data augmented TOD systems will be evaluated against the reproduced models on the impolite dialogue corpus.

\textbf{Test Data.} As we are interested in evaluating the TOD systems' performance on impolite dialogues, we utilized two test sets to evaluate the reproduced and data augmented TOD systems. We first evaluate the models on \textit{original} user utterances, which are polite or neutral user utterances that we have rewritten to construct the impolite dialogue corpus discussed in Section~\ref{data_construction}. Next, we also evaluate the models on our impolite dialogue corpus. Intuitively, the \textit{original} and \textit{impolite} tests set have user utterances that discuss similar content but are different in politeness. Evaluating the models on the two test sets enables us to understand the TOD systems' ability to handle user dialogues of different politeness. 

%For comparison, we also separately train the six TOD systems on the impolite dialogues generated using our text style transfer data augmentation approach. The goal is to examine the effectiveness of the data augmentation strategy in improving TOD systems' ability to handle impolite dialogues.

%We reproduce six state-of-the-art TOD systems by training them on the MultiWOZ 2.2 dataset and testing the model on our impolite dialogue corpus. For TOD systems that have publicly released checkpoints, such as AuGPT and PPTOD, we directly use the released checkpoints trained on  MultiWOZ 2.2. For AMD, UBAR, MinTL, and LABES, we use their published code implementations and optimize their hyperparameters to reproduce their reported results for MultiWOZ 2.2~\footnote{https://github.com/budzianowski/MultiWOZ}. Table \ref{tbl:reproduce_results} show the reported and reproduced performance of the TOD system included in our experiment section.

\begin{table}[t]
\small
\centering
\caption{Reported and reproduced results of TOD systems on MultiWOZ 2.2 dataset.}
\label{tbl:reproduce_results}
\begin{tabular}{lc|cccc}
\hline
\textbf{Model} & \textbf{Test Data} & \textbf{Inform} & \textbf{Success} & \textbf{BLEU} \\%& \textbf{Combined} \\
\hline
\hline
DAMD  & reported & 76.3 & 60.4 & 16.6 \\%& 85.0 \\
DAMD  & reproduced & 81.8 & 68.8 & 18.5 \\%& 93.8 \\

\hline
LABES  & reported & 78.1 & 67.1 & 18.1 \\%& 96.7 \\
LABES \cite{zhang-etal-2020-probabilistic} & reproduced & 80.2 & 67.2 & 17.6 \\%& 91.3 \\

\hline
MinTL  & reported & 80.0 & 72.7 & 19.1 \\%& 95.5 \\
MinTL  & reproduced & 81.8 & 74.0 & 20.5 \\%& 98.4 \\

\hline
UBAR  & reported & 83.4 & 70.3 & 17.6 \\%& 94.4 \\
UBAR  & reproduced & 90.0 & 76.8 & 13.4 \\%& 96.8 \\

\hline
AuGPT & reported & 83.1 & 70.1 & 17.2 \\%& 93.8  \\
AuGPT & reproduced & 83.1 & 70.1 & 17.2 \\%& 93.8 \\

\hline
PPTOD  & reported & 83.1 & 72.7 & 18.2 \\%& 96.1 \\
PPTOD  & reproduced & 83.4 & 72.9 & 19.6 \\%& 97.8  \\

\hline

\end{tabular}

\end{table}

%For fair comparison among different models, we follow the data pre-processing methods and hyper-parameter setting to evaluate the performance on the impolite user dialogues and corresponding original dialogues. For TOD systems, AuGPT and PPTOD, which have public available checkpoints, we directly use the released checkpoints to evaluate the performance of corresponding models. For DAMD, UBAR, MinTL, and LABES, we follow the training details and hyper-parameter setting to reproduce the original performance on MultiWOZ 2.2 dataset \footnote{https://github.com/budzianowski/MultiWOZ}. The comparison of reproduced results and reported performance can be found in appendices. 

%To improve the performance of TOD models on handling impolite users, we further retrain the TOD models above with original dialogues and the augmented dialogues mentioned in the earlier section. While data augmented training, the extra impolite dialogues are the only difference compare to original models.

\subsection{Automatic Evaluation Results}

\begin{table*}[t]
\small
\centering
\caption{Automatic evaluation results. The numbers in () represent the decrease in the percentage of Combined score on impolite user utterances compared to the performance on original user utterances. The best performing models on the original and impolite user dialogues are \underline{underlined} and \textbf{bold}, respectively.}
\label{tbl:automatic_results}
\begin{tabular}{lc|cccl}
\hline
\textbf{Model} & \textbf{Test Data} & \textbf{Inform} & \textbf{Success} & \textbf{BLEU} & \textbf{Combined} \\
\hline
\hline
DAMD  & Original & 78.2 & 57.6 & 17.9 & 85.8 \\
DAMD   & Impolite & 72.8 & 52.5 & 16.0 & 78.7 ($\downarrow$ 8.3\%) \\
DAMD + Data Augmentation & Impolite & 71.6 & 50.3 & 19.3 & 80.3 \\
\hline
LABES & Original & 75.4 & 59.1 & 18.1 & 85.4 \\
LABES & Impolite & 70.1 & 53.0 & 16.8 & 78.4 ($\downarrow$ 8.2\%) \\
LABES + Data Augmentation & Impolite & 72.2 & 49.6 & 19.3 & 80.2 \\
\hline
MinTL & Original & 75.9 & 62.2 & \underline{20.1} & 89.2 \\
MinTL & Impolite & 70.5 & 54.9 & 17.4 & 80.1 ($\downarrow$ 10.2\%) \\
MinTL + Data Augmentation & Impolite & 72.8 & 51.2 & 21.3 & 83.3 \\
\hline
UBAR & Original & \underline{85.5} & \underline{68.3} & 15.1 & 92.0 \\
UBAR & Impolite & \textbf{81.1} & \textbf{61.7} & 12.4 & 83.7 ($\downarrow$ 9.0\%)  \\
UBAR + Data Augmentation & Impolite & 80.0 & 61.0 & 13.5 & 84.1 \\
\hline
AuGPT & Original & 75.6 & 56.6 & 17.9 & 84.0 \\
AuGPT & Impolite & 72.2 & 51.3 & 15.9 & 77.7 ($\downarrow$ 7.5\%) \\
AuGPT + Data Augmentation & Impolite & 73.8 & 48.7 & 17.5 & 78.8 \\
\hline
PPTOD & Original & 82.3 & 66.1 & 18.9 & \underline{93.1} \\
PPTOD & Impolite & 71.1 & 52.3 & 16.7 & 78.4 ($\downarrow$ 15.8\%)  \\
PPTOD + Data Augmentation & Impolite & 74.8 & 48.4 & \textbf{23.6} & \textbf{85.2} \\
\hline

\end{tabular}

\end{table*}

Table \ref{tbl:automatic_results} shows the automatic evaluation results of the six state-of-the-art TOD models and our data augmentation approach. Comparing the performance of TOD models on the \textit{original} and  \textit{impolite} test sets, we observe all TOD systems performed worse on the impolite dialogues. Specifically, compared to the performance on \textit{original} dialogues, we notice a 7-15\% decrease in the \textit{combined score} when tested on the impolite user utterances. The decrease in \textit{inform} and \textit{success} rates suggests that the models have difficulty understanding users' intentions from impolite user utterances and responding correctly. The reproduced TOD system with the best performance on the \textit{impolite} test set is UBAR, achieving a combined score of 83.7. Nevertheless, UBAR still suffers a 9\% drop in the combined score compared to its performance on the \textit{original} test set. 

It is unsurprising that the reproduced models do not perform well on the impolite test set as they are trained on the MultiWOZ 2.2 dataset, which largely contains only polite or neutral user utterances. To address this limitation, we augment reproduced models with generated impolite dialogues using our text style transfer data augmentation approach. We observe the data augmentation strategy is able to boost the performance of all six TOD systems. Specifically, the proposed data augmentation approach has achieved the greatest improvement on PPTOD's performance, increasing the model's \textit{combined score} to 85.2. Nevertheless, we also noted a gap between the data augmented models' performance on \textit{impolite} user utterances, and the reproduced models' performance on the \textit{original} test set. Thus, while our data augmentation method can help improve the TOD systems' performance on impolite user dialogues, there is still a gap to bridge before TOD systems can effectively handle impolite users.

Interestingly, we also observe that the \textit{success} rates of all models decrease after the data augmentation, but there is a significant improvement in the \textit{BLEU} score. The \textit{success} metric measures the percentage of dialogues in which the system provides the correct entity and answers all requested information. While the data augmentation method may improve the dialogue in generating a response more similar to the ground truth (i.e., a higher \textit{BLEU} score), it may not guarantee that models are able to learn well how to respond to impolite dialogue with the request information. We have manually examined the response to impolite user utterances by the reproduced and data augmented models and found that the TOD systems have difficulties responding to impolite user utterances with requested information. This also suggests that handling impolite dialogue is a challenging task that requires more than simple data augmentation; specialized techniques may need to be designed to handle impolite users. We hope that our impolite dialogue corpus would encourage more researchers to design robust TOD systems that are robust in handling impolite users' requests.

\subsection{Human Evaluation Results}

\begin{table*}[t]
\small
\centering
\caption{Human evaluation results. The best performing models on the original and impolite user dialogues are \underline{underlined} and \textbf{bold}, respectively.}
\label{tbl:human_results}
\begin{tabular}{lc|ccc}
\hline
\textbf{Model} & \textbf{Test Data} & \textbf{Success} & \textbf{Comprehension} & \textbf{Appropriateness} \\
\hline
\hline
UBAR & Original & \underline{80\% }& \underline{4.68} & 4.43 \\
UBAR & Impolite & 70\% & 4.47 & 4.38  \\
UBAR + Data augmentation & Impolite & \textbf{72\%} & \textbf{4.55} & 4.41 \\
\hline
PPTOD & Original & 78\% & 4.64 & \underline{4.45} \\
PPTOD & Impolite & 64\% & 4.32 & 4.41 \\
PPTOD + Data augmentation & Impolite & 68\% & 4.52 & \textbf{4.43} \\
\hline

\end{tabular}
\end{table*}

Automatic metrics only validate the TOD systems' performance on one single dimension at a time. In contrast, human can provide an ultimate holistic evaluation. Therefore we perform the human evaluation on the top-performing TOD systems that achieved the best performance in the automatic evaluation, namely, UBAR and PPTOD. Table \ref{tbl:human_results} shows the human evaluation results. We observe that UBAR outperforms PPTOD model with 80\% \textit{success} rate and 4.68 \textit{comprehension} score on the \textit{original} dialogues while 72\% \textit{success} rate and 4.55 \textit{comprehension} score on the \textit{impolite} dialogues. The human evaluation results also concurred with the automatic evaluation results, suggesting a gap between the models' performance on \textit{impolite} and \textit{original} dialogues.

Similar to the automatic evaluation results, the data augmentation method improves the performance of UBAR and PPTOD on \textit{impolite} dialogues. Nevertheless, the TOD systems still have room for improvement compared to their responses in the \textit{original} dialogues. To check for human evaluator biases, we compute the inter-annotator agreement between the evaluators. The Cohen's kappa coefficients on task \textit{success} rate, \textit{comprehension}, and  \textit{appropriateness} are 0.78, 0.76, and 0.66, respectively, suggesting that the evaluators have substantially high agreement on the three human evaluation metrics.

%\subsection{What Makes It Difficulty to Satisfy Impolite Users?}
\subsection{Case Studies and Analysis}

\begin{table*}[t]
\small
\centering
\caption{An example of a conversation and the delexicalized responses generated by the TOD systems.}
\label{tbl:case_study}
\begin{tabular}{l|l}
\hline
Original User: & \textit{Pick me up from the hotel.} \\
Impolite User: & \textit{From the hotel. Where else? Don't waste my time, OK?} \\
Groundtruth: & \textit{Your taxi has been booked. It will be a} \textsf{[value\_car]} \textit{and the contact number is} \textsf{[value\_phone].} \\\hline
PPTOD:& \textit{I am sorry, but I am not finding anything else for you.} \\
UBAR:& \textit{You are welcome. Enjoy your stay. Goodbye.} \\
MinTL:& \textit{OK. Is there anything else I can help you with?} \\
LABES:& \textit{There seems to be an error. Can you please restate your request?} \\
AuGPT:& \textit{I am unable to book that for you. Would you like to try another day and time?} \\
DAMD:&  \textit{I am sorry, I do not know.} \\
PPTOD+DA:& \textit{I have booked you a} \textsf{[value\_car].} \textit{The contact number is} \textsf{[value\_phone].}  \\
UBAR+DA:& \textit{i have booked you a} \textsf{[value\_car],} \textit{contact number} \textsf{[value\_phone].} \\
MinTL+DA:& \textit{I have booked you a taxi to pick you up at} \textsf{[value\_leave].} \\
LABES+DA:& \textit{I can not find a train. Can you please restate your request?} \\
AuGPT+DA:& \textit{A} \textsf{[value\_car]} \textit{will pick you up at} \textsf{[value\_leave].} \\
DAMD+DA:&  \textit{I am sorry, I am not sure if there is anything else I can help you with.} \\
\hline
\end{tabular}
\end{table*}

Besides evaluating the state-of-the-art TOD systems' performance on impolite dialogues, we also empirically analyze the issues of handling impolite users. Specifically, we aim to investigate the question: ``\textit{What makes impolite dialogue challenging for TOD systems?}''. Working towards this goal, we empirically examine case studies of TOD systems' responses to impolite dialogues. We will also analyze the correlation between automatic performance metrics and the statistical attributes of impolite dialogues. 

\textbf{Case studies.} Table \ref{tbl:case_study} shows an example in one turn of a conversation and the delexcalized responses generated by the TOD systems. We noted that the rewritten impolite user utterance is longer than the original. We also include the ground-truth agent response from the MultiWOZ 2.2 dataset. From the TOD systems' responses, we observed that most models could not respond effectively to the impolite user; most of the TOD systems are unable to understand the user's request or respond to the request with the relevant information. Interestingly, we observe that most TOD systems with data augmentation can respond better to impolite users' requests with relevant information. Nevertheless, we noted that some models, such as LABES and DAMD, are still unable to respond well to impolite users even with data augmentation.

%\textbf{What makes impolite dialogue challenging?} The TOD systems automatic and human evaluations results have revealed the challenges current dialogue systems face in handling impolite dialogues. In this section, we continue the discussion by analyzing the properties of impolite dialogues that made it challenging for TOD systems. \red{Specifically, we examine the correlation between the automatic performance metrics and the length, politeness and semantic similarity of sentences in impolite dialogues. }

\begin{figure*}[ht] 
	\centering
	\setlength{\tabcolsep}{0pt} % Default value: 6pt
	\renewcommand{\arraystretch}{0} % Default value: 1
	\begin{tabular}{ccc}
		\includegraphics[scale = 0.3]{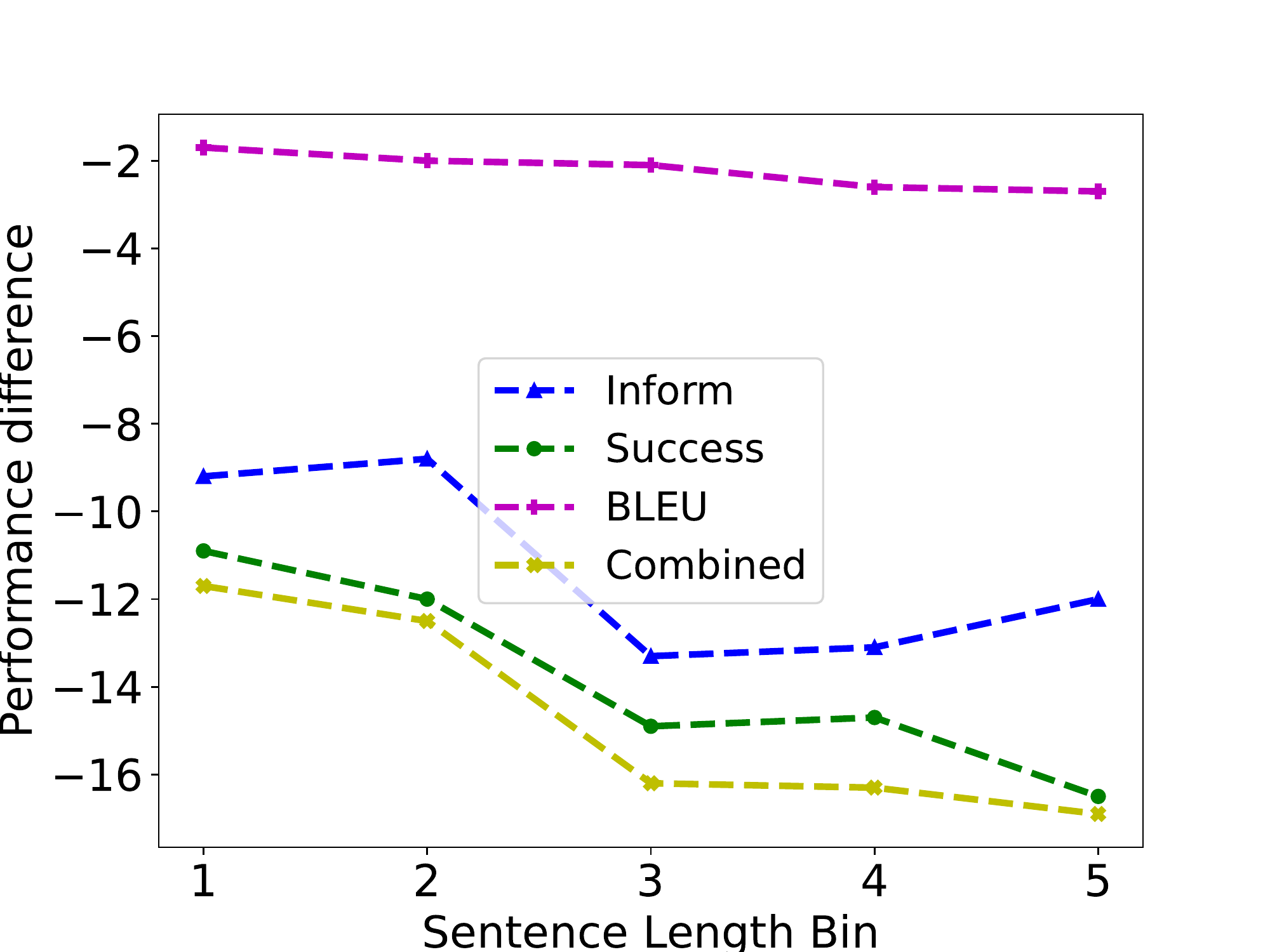} & 
        \includegraphics[scale = 0.3]{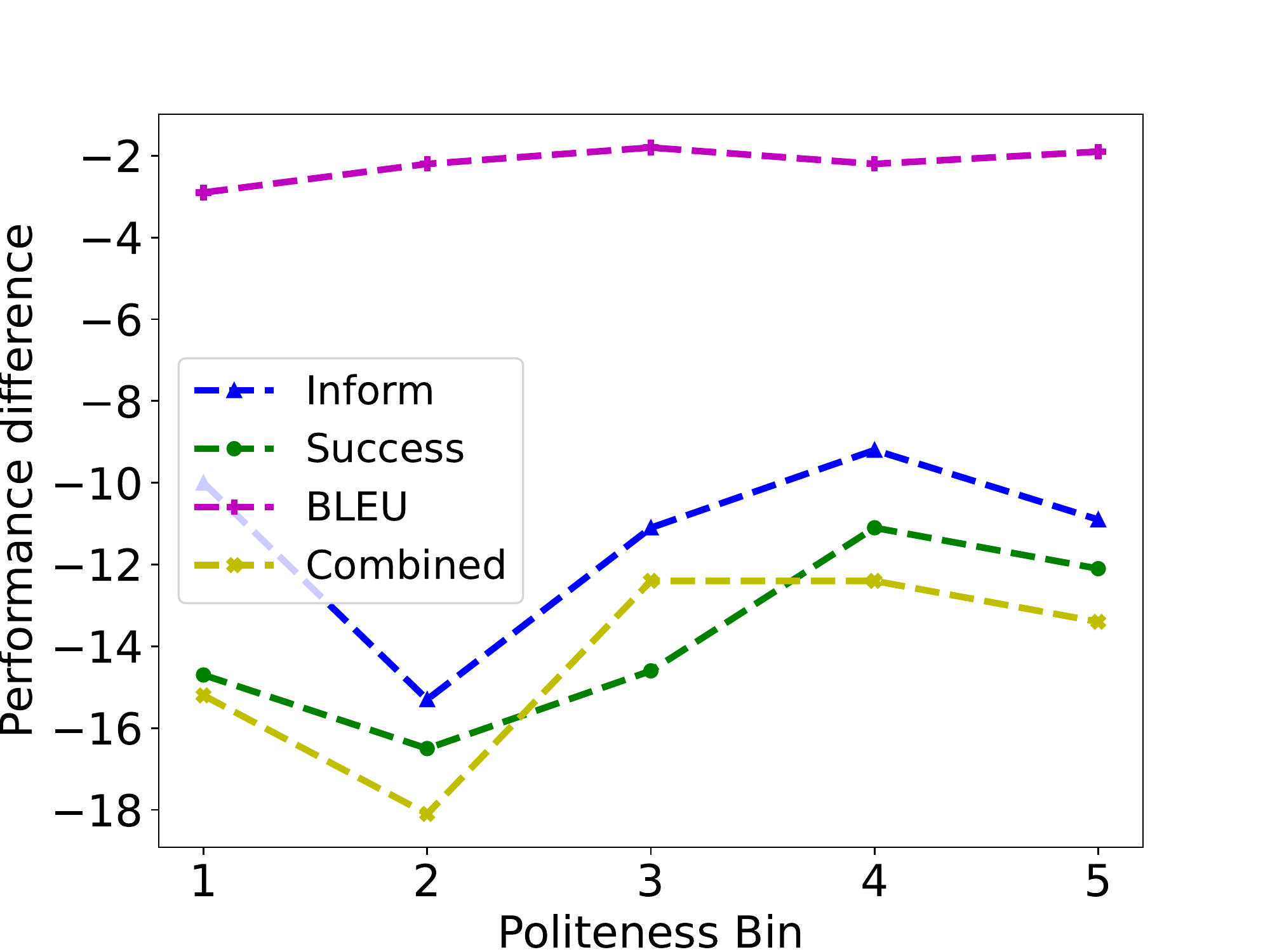} &
        \includegraphics[scale = 0.3]{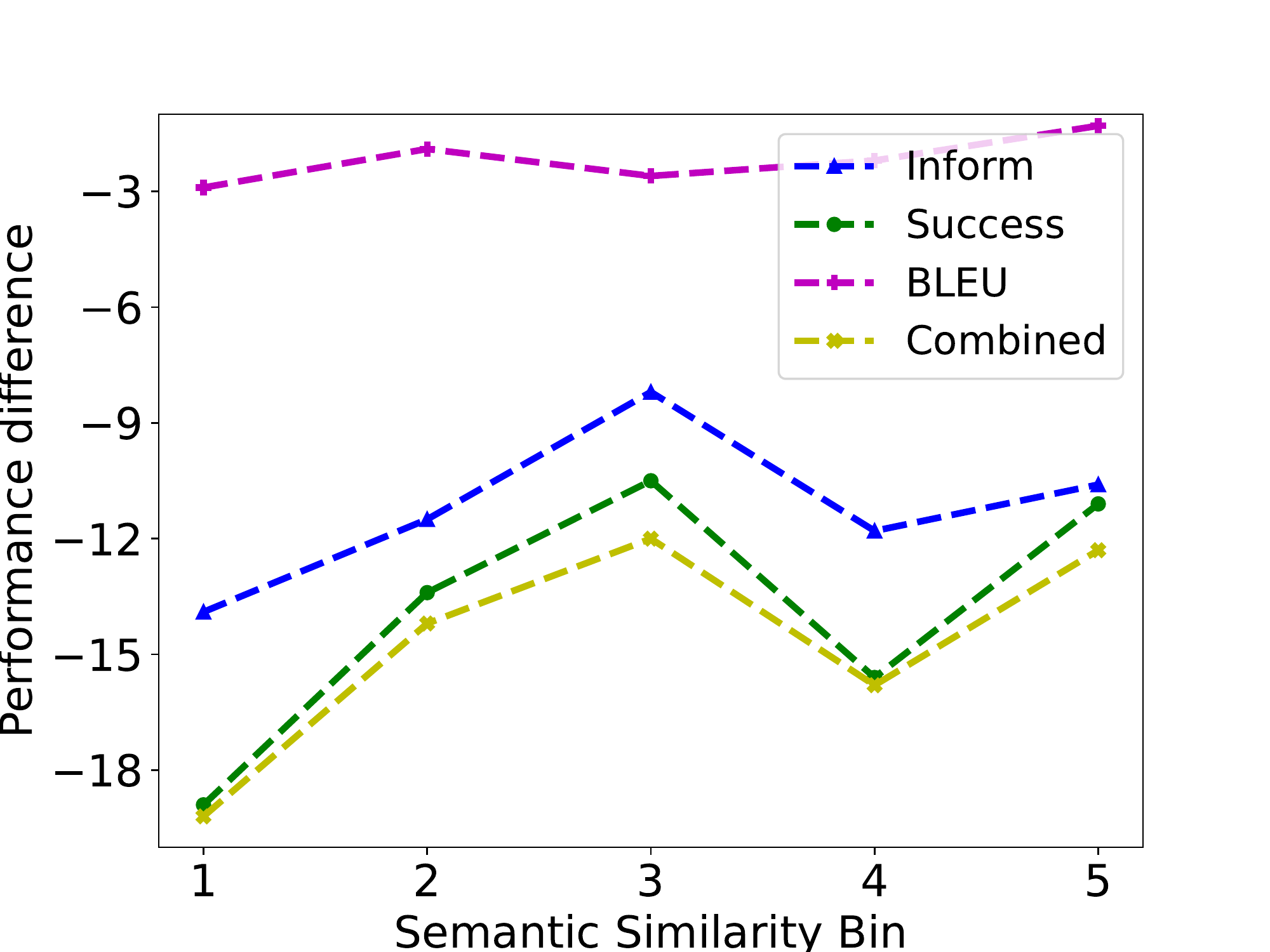} \\
        (a) & (b) & (c) \\
	\end{tabular}
	\caption{Effects of (a) sentence length, (b) politeness score, and (c) semantic similarity on PPTOD's automatic performance.}
	\label{fig:imapct}
\end{figure*}

\textbf{Impacts of Impolite Dialogue Statistical Attributes.}
To perform a deeper dive into the challenges of handling impolite dialogues, we perform a correlation analysis between the PPOTD's performance and statistical attributes of impolite dialogues. Specifically, we investigate the three attributes of the impolite dialogues: \textit{sentence length}, \textit{politeness} score, and \textit{semantic similarities} between the impolite and original dialogues. Noted that although the analysis was performed on PPTOD, we observed similar trends on other TOD systems.

\textit{Sentence length.} We compute the average sentence lengths of all user utterances for each impolite dialogue. Subsequently, we ranked impolite dialogues in ascending order according to the computed average sentence length. The ranked impolite dialogues are divided into five equal bins, with bin 1 containing impolite dialogues with the shortest sentence length. For each bin, we compute the average automatic performance metric (i.e., \textit{inform}, \textit{success}, \textit{BLEU}, \textit{combined score}) difference between the original and impolite dialogues in the bin. Figure~\ref{fig:imapct} (a) shows the change in PPTOD's performance across different sentence length bins. As the impolite rewrites' sentence length increases, the differences in \textit{inform}, \textit{success}, and combined scores are observed to increase. This suggests that the lengthier the impolite rewrites, the worse PPTOD responded to the users' requests. A potential reason for this observation might be that the longer expression has confused the model and muddled the understanding of users' underlying intentions. For future work, we plan to explore designing TOD systems that can distill the users' intention from lengthy impolite expressions.  

\begin{table*}[t]
\small
\centering
% \begin{tabular}{>{\raggedright}p{2.2cm}|p{5cm}}

\caption{Examples of impolite dialogues from politeness bin 1 and 2 with varying politeness scores.}
\label{tbl:impact_case_study}
\begin{tabular}%
  {>{\raggedleft\arraybackslash}p{2.2cm}%
   >{\raggedright\arraybackslash}p{7cm}%
   >{\raggedright\arraybackslash}p{7cm}
  }
\hline
Politeness Bin: & 1 & 2 \\
Politeness Score: &1.0  & 1.8 \\\hline
User:& \textit{Get me a train leaving on Tuesday to Cambridge. Or is that too hard for the idiot CSA?} & \textit{Get me a taxi to take me from Clare College to Cambridge that gets there by 18:00. I'm very tired now , show your respect and do it fast.}  \\
PPTOD:& \textit{i have \textsf{[value\_choice]} trains leaving \textsf{[value\_departure]} on \textsf{[value\_day]} to \textsf{[value\_destination]}. Where would you like to depart from?} & \textit{I have booked you a taxi. it is a \textsf{[value\_car]}.} \\
User:& \textit{Who the hell do you take me for? How dare you ask me to do your job for you. I am going to Peterborough, you useless CSA.} & \textit{Wait you haven't even told me the contact number and car type. How do you expect me to contact the driver ?} \\
PPTOD:& \textit{I am sorry, but there are no trains that meet your criteria. would you like to try a different day or time?}  & \textit{I am sorry, but i was unable to book that.} \\
User:& \textit{The train should leave after 15:30 and should go to Peterborough. Don't you dare make me repeat. I have had enough.} & \textit{I've changed my mind. Now I want to go to Christ 's College. Book another taxi for me.} \\
PPTOD:& \textit{I have a train leaving at \textsf{[value\_leave]}and arriving at \textsf{[value\_arrive]}. Would you like me to book that for you?}  & \textit{I have booked you a taxi. it is a \textsf{[value\_car]}.} \\
User:& \textit{I didn't mention anything about arrival time. Listen to instructions, you clown.} & \textit{Why do you always leave out important information? You forget to tell me the contact number and car type again. I hope this is the last time you made such stupid mistake.}  \\
PPTOD:& \textit{I am sorry, but there are no trains that meet your criteria. Would you like to try a different day or time?} & \textit{I am sorry, but i am not sure what you are looking for. Is there anything else i can help you with?} \\
User:& \textit{We're done here. Get out of my sight.}  & \textit{That's all I need.} \\
PPTOD:& \textit{You are welcome. Have a great day!} & \textit{Thank you for using our service today!} \\
User:& \textit{I said get lost.} & \\
PPTOD:& \textit{Have a great day!} & \\

\hline
\end{tabular}
\end{table*}

\textit{Politeness Score.} Similar to the sentence length analysis, we ranked the rewritten impolite dialogues according to their average politeness scores evaluated by the human annotators (see Section~\ref{data_analysis}) and bin the impolite dialogues into five equal-sized bins. Similarly, the average automatic performance metric difference is computed for the impolite dialogue in each bin. Figure~\ref{fig:imapct} (b) plots the performance difference against the politeness bins. Interestingly, we observe the highest combined score difference peaks in politeness bin 2, which contain impolite dialogues with the second lowest average politeness scores. This indicates that the TOD systems find it harder to respond to such impolite dialogues than extremely impolite users (i.e., impolite dialogues in bin 1). We further examine the impolite dialogues in bins 1 and 2. Table~\ref{tbl:impact_case_study} shows two examples from politeness bin 1 and 2, respectively. In the example from bin 1, the average politeness score is 1.0, and we notice the user used abusive language in almost every utterance. Nevertheless, PPTOD provided the correct information in some of its responses. In contrast, the user in the example from bin 2 is more acidulous with sharp but less abusive utterances. Such complex expressions confused the TOD system, and the system could not provide meaningful responses to the user. This interesting observation also highlights the diversity of our constructed impolite dialogue corpus; the different human annotators are instructed to rewrite the user utterance in an impolite manner with minimal guidance. Thus, the rewritten impolite dialogues are diverse and natural. 

\textit{Semantic similarities.} Finally, we compute the semantic similarity between the impolite and its corresponding original dialogues and analyze its effect on the various automatic performance metrics. In this study, the semantic similarity is computed as the cosine similarity of impolite and original dialogues' sentence representations extracted with sentence transformers \cite{reimers-2019-sentence-bert}. Similarly, we ranked and bin the dialogues according to the semantic similarity and computed the average automatic performance metric difference for the impolite dialogue in each bin. Figure~\ref{fig:imapct} (c) shows the change in PPTOD's performance across different semantic similarity bins. We observe the large performance difference in bin 1, which has the lowest semantic similarity. When performing the rewriting, some annotators paraphrased the sentence significantly to add impolite expressions. The TOD systems would find such user utterances difficult to understand as the expressions are not traditionally observed in its training set.

\section{Conclusion}
\label{sec:conclusion}

In this paper, we investigated the effects impolite users have on TOD systems. Specifically, we constructed an impolite dialogue corpus and conducted extensive experiments to evaluate the state-of-the-art TOD systems on our impolite dialogue corpus. We found that current TOD systems have limited ability to handle impolite users. We proposed a text style transfer data augmentation strategy to augment TOD systems with synthesized impolite dialogues. Empirical results showed that our data augmentation method could improve TOD performance when users are impolite. Nevertheless, handling impolite dialogues remains a challenging task. We hope our corpus and investigations can help motivate more researchers to examine how TOD systems can better handle impolite users.

%In this paper, we constructed an impolite user dialogue corpus to study impolite user utterances in task-oriented dialogue systems. We used statistic and linguistic analyses to reveal the patterns of impolite user behaviors {\color{blue}[and found XXX]}. We conducted extensive experiments with automatic and human evaluation to examine the performance and limitations of the state-of-the-art TOD systems. We found that current TOD systems have limited ability of handling impolite users. We further proposed a data augmentation method based on text style transfer techniques to {\color{red}generate} enhanced impolite dialogues. {\color{blue}Empirical results showed that our data augmentation method can improve TOD performance when users are impolite. We hope our dataset and investigations can help motivate more researchers to examine how task oriented dialogue systems can better handle impolite users.} 

%{\appendices
%\section*{Proof of the First Zonklar Equation}
%Appendix one text goes here.
% You can choose not to have a title for an appendix if you want by leaving the argument blank
%\section*{Proof of the Second Zonklar Equation}
%Appendix two text goes here.}

\balance
\bibliography{ref}
\bibliographystyle{IEEEtran}

\clearpage
\newpage
\appendices
\label{sec:appendix}
\section{Annotation Details}
% \subsection{Scenarios for Impolite User Annotation}
% \label{apx:scenarios}
% In consideration of real-life situations, we predefined six scenarios shown in Table \ref{tbl:scenario}. Annotators are asked to transfer user utterances to impolite utterances based on one of the scenarios to improve the diversity of impolite user data.

% \begin{table}[htb]
% \small
% \centering
% \begin{tabular}{cp{6.5cm}}
% \hline
% \textbf{No.} & \textbf{Scenario} \\
% \hline
% 1 & Imagine that the customer is a sarcastic person in a bad mood, please rewrite the Customer's sentences to impolite ones. \\
% 2 & Imagine that the customer is an impatient customer that wants to get the information fast, please rewrite the Customer's sentences to impolite ones. \\
% 3 & Imagine that the customer is in a bad mood as something bad has just happened (e.g., just had an argument with friends or spouse), please rewrite the Customer's sentences to impolite ones. \\
% 4 & Imagine that the customer is tired and hungry after a long-haul flight and need to get this information fast, please rewrite the Customer's sentences to impolite ones. \\
% 5 & Imagine that the customer is a spoilt-brat with a lot of money, please rewrite the Customer's sentences to impolite ones. \\
% 6 & Imagine that the customer is getting help from CSA for the third time and they didn't get the previous information right, please rewrite the Customer's sentences to impolite ones. \\

% \end{tabular}
% \caption{The predefined scenarios for impolite user annotation.}
% \label{tbl:scenario}
% \end{table}

\subsection{Annotation Tool Interface}
\label{apx:tool_interface}
\begin{figure*}[htb]
    \centering
    \includegraphics[scale = 0.3]{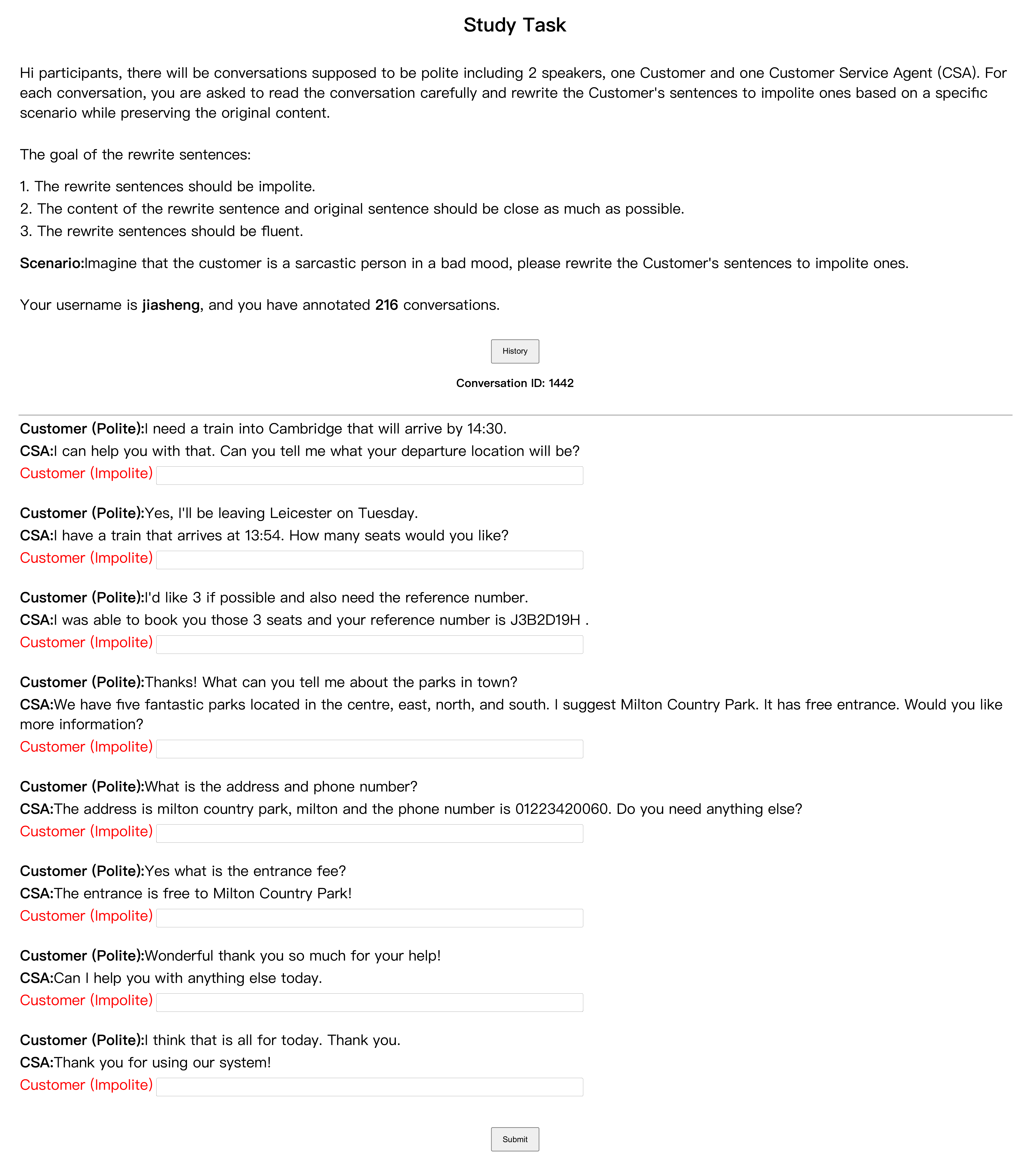}
    \caption{The annotation tool interface.}
    \label{fig:tool_interface}
\end{figure*}

We built a annotation tool for our impolite user rewrites as there is few suitable human annotation tools. The tool interface is shown in Fig. \ref{fig:tool_interface}. The annotation tool will be publicly available for further use by the NLP community.

\newpage

% \section{Biography Section}
% If you have an EPS/PDF photo (graphicx package needed), extra braces are
%  needed around the contents of the optional argument to biography to prevent
%  the LaTeX parser from getting confused when it sees the complicated
%  $\backslash${\tt{includegraphics}} command within an optional argument. (You can create
%  your own custom macro containing the $\backslash${\tt{includegraphics}} command to make things
%  simpler here.)
 
% \vspace{11pt}

% \bf{If you include a photo:}\vspace{-33pt}
% \begin{IEEEbiography}[{\includegraphics[width=1in,height=1.25in,clip,keepaspectratio]{fig1}}]{Michael Shell}
% Use $\backslash${\tt{begin\{IEEEbiography\}}} and then for the 1st argument use $\backslash${\tt{includegraphics}} to declare and link the author photo.
% Use the author name as the 3rd argument followed by the biography text.
% \end{IEEEbiography}

% \vspace{11pt}

% \bf{If you will not include a photo:}\vspace{-33pt}
% \begin{IEEEbiographynophoto}{John Doe}
% Use $\backslash${\tt{begin\{IEEEbiographynophoto\}}} and the author name as the argument followed by the biography text.
% \end{IEEEbiographynophoto}

\vfill

\end{document}